%% file: main.tex
\documentclass[runningheads]{llncs}
\usepackage[T1]{fontenc}
\usepackage{graphicx}
\usepackage{color}
\input{util/packages}
\input{util/macros}

\input{util/lstsettings}
\input{plots/plotting_methods.tex}
\begin{document}
\title{Easy-to-Use Shielding for Reinforcement Learning}
\author{Stefan Pranger\inst{1}\orcidID{0009-0000-6011-9925} \and
Bettina Könighofer\inst{1}\orcidID{0000-0001-5183-5452}}
\authorrunning{S. Pranger and B. Könighofer}
\institute{Institute of Information Security, Graz University of Technology
\email{stefan.pranger@tugraz.at\\bettina.koenighofer@tugraz.at}}
\maketitle              
\begin{abstract}
  \input{00_Abstract}
\end{abstract}
\section{Introduction}
\input{01_Introduction}
\input{02_RelatedWork}
\section{Background}\label{sec:background}
\input{03_Background}
\section{Seamless Integration of Shielding using \tempestpy}\label{sec:tempestpy}
\input{04a_Method_Synthesis.tex}
\input{04b_Method_Application_RL.tex}
\section{\minigridsafe}\label{sec:minigridsafe}
\input{06_MinigridSafe}
\section{Easy Shielding in \minigridsafe}\label{sec:experiments}
\input{07_Shielding_With_Tempestpy}

\section{Tool Architecture}
\input{08_Tool_Architecture}
\section{Conclusion}\label{sec:conclusion}
\input{09_Conclusion}
\bibliographystyle{splncs04}
\bibliography{main}
\end{document}

%% file: util/packages.tex
\usepackage{graphicx}
\usepackage[T1]{fontenc}
\usepackage[utf8]{inputenc}
\usepackage[misc]{ifsym}
\usepackage{tikz}
\usepackage{url}
\usepackage{amsmath,amssymb,amsfonts}
\usetikzlibrary{arrows,arrows.meta,shapes.gates.logic.US,shapes.gates.logic.IEC,calc,decorations.markings,fit,positioning,shapes.symbols,backgrounds}
\usepackage{siunitx}
\usepackage{pgfplots}
\usepackage{paralist}
\pgfplotsset{compat=1.18}
\usepgfplotslibrary{statistics}
\usepgfplotslibrary{fillbetween}
\usepackage[acronym]{glossaries}
\usepackage{multirow}
\usepackage{enumitem}
\usepackage{todonotes}
\usepackage{xspace}
\usepackage{blindtext}
\usepackage[british]{babel}
\usepackage{lipsum}
\setlipsum{
  par-before = \begingroup\color{red},
  par-after = \endgroup
}
\usepackage{gensymb}
\usepackage{caption}
\usepackage[aboveskip=-1pt, belowskip=-1pt]{subcaption}
\usepackage{listings}
\usepackage{xcolor}
\usepackage{scalefnt}
\usepackage{booktabs}

%% file: util/macros.tex
\newcommand{\Dist}{\mathit{Dist}}

\newcommand{\storm}{\textsc{Storm}\xspace}
\newcommand{\stormpy}{\textsc{stormpy}\xspace}
\newcommand{\tempest}{\textsc{Tempest}\xspace}
\newcommand{\prism}{\textsc{Prism}\xspace}
\newcommand{\prismgames}{\textsc{Prism-games}\xspace}

\newcommand{\tempestpy}{\textsc{tempestpy}\xspace}

\newcommand{\minigrid}{\textsc{Minigrid}\xspace}

\newcommand{\minigridsafe}
{\textsc{Minigrid}\raisedsmall{\textsc{Safe}}\xspace}
\def\raisedsmall#1{\raise.2em\hbox{$\fontsize{7pt}{0pt}\selectfont#1$}}

\def\raisedsmaller#1{\raise.2em\hbox{$\fontsize{6pt}{0pt}\selectfont#1$}}

\newcommand{\mdp}{\ensuremath{\mathcal{M}}\xspace}
\newcommand{\smg}{\ensuremath{\mathcal{G}}\xspace}

\newcommand{\risk}{\ensuremath{\mathit{risk}}}
\newcommand{\presh}{\ensuremath{\sigma_{pre}}}
\newcommand{\postsh}{\ensuremath{\sigma_{post}}}
\newcommand{\rlTrans}{\ensuremath{\tau}}
\newcommand{\thresh}{\ensuremath{\delta_{\varphi}}}
\newcommand{\horiz}{\ensuremath{h}}
\newcommand{\coali}{\ensuremath{\Pi'}}
\renewcommand{\phi}{\varphi} 
\newcommand{\mdpobjective}{\ensuremath{\langle \varphi,\thresh\rangle}}

\newcommand{\trans}{\mathcal{P}}
\newcommand{\states}{\ensuremath{\mathcal{S}}}
\newcommand{\Act}{\mathcal{A}}
\newcommand{\rewards}{\mathcal{R}}
\newcommand{\rewFunction}{\rewards}

\newcommand{\players}{\ensuremath{\Pi}}

\newcommand{\ph}[2][4pt]{\vspace{#1}\noindent\textbf{#2.}}

\newcommand{\shieldIcon}{\includegraphics[scale=0.015]{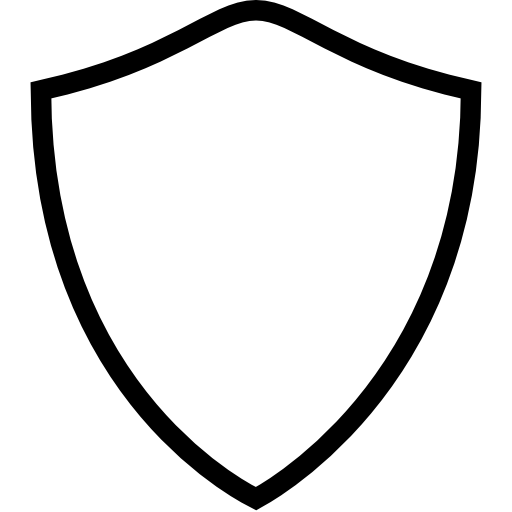}}

%% file: util/lstsettings.tex
\definecolor{codebg}{RGB}{248,248,248}
\definecolor{keywordcolor}{RGB}{0,0,180}
\definecolor{commentcolor}{RGB}{100,100,100}
\definecolor{stringcolor}{RGB}{160,0,0}
\definecolor{identcolor}{RGB}{0,100,100}
\lstdefinestyle{tempest}{
  language=Python,
  backgroundcolor=\color{codebg},
  basicstyle=\ttfamily\small,
  keywordstyle=\color{keywordcolor}\bfseries,
  commentstyle=\color{commentcolor}\itshape,
  stringstyle=\color{stringcolor},
  emphstyle=\color{identcolor}\bfseries,
  emph={Shield, MaskingConfig, ShieldWrapper, MaskResult},
  breaklines=true,
  breakatwhitespace=true,
  showstringspaces=false,
  frame=single,
  framerule=0.4pt,
  rulecolor=\color{gray!50},
  numbers=left,
  numberstyle=\tiny\color{gray},
  numbersep=6pt,
  xleftmargin=2em,
  captionpos=b,
  belowskip=0.8em,
  aboveskip=0.8em,
}

%% file: plots/plotting_methods.tex
\newcommand{\lavaPlot}[2]{
    \addplot [mark=none, color=#2, solid] table [y=info/sum_ran_into_lava_avg, x=time/total_timesteps]{#1};
    \addplot [mark=none, name path=lava_max, color=#2!20, draw opacity=0.1,forget plot] table [y=info/sum_ran_into_lava_max, x=time/total_timesteps]{#1};
    \addplot [mark=none, name path=lava_min, color=#2!20, draw opacity=0.1,forget plot] table [y=info/sum_ran_into_lava_min, x=time/total_timesteps]{#1};
    \addplot[#2!20, opacity=0.4,forget plot] fill between[of=lava_min and lava_max];
}

\newcommand{\rewardPlot}[2]{
    \addplot [mark=none, color=#2, solid] table [y=rollout/ep_rew_mean_avg, x=time/total_timesteps]{#1};
    \addplot [mark=none, name path=rew_max, color=#2!20, draw opacity=0.1] table [y=rollout/ep_rew_mean_max, x=time/total_timesteps]{#1};
    \addplot [mark=none, name path=rew_min, color=#2!20, draw opacity=0.1] table [y=rollout/ep_rew_mean_min, x=time/total_timesteps]{#1};
    \addplot[#2!20, opacity=0.4] fill between[of=rew_min and rew_max];
}
\newcommand{\collisionPlot}[2]{
    \addplot [mark=none, color=#2, solid] table [y=info/sum_collision_avg, x=time/total_timesteps]{#1};
    \addplot [mark=none, name path=col_max, color=#2!20, draw opacity=0.1,forget plot] table [y=info/sum_collision_max, x=time/total_timesteps]{#1};
    \addplot [mark=none, name path=col_min, color=#2!20, draw opacity=0.1,forget plot] table [y=info/sum_collision_min, x=time/total_timesteps,forget plot]{#1};
    \addplot[#2!20, opacity=0.4,forget plot] fill between[of=col_min and col_max];
}
\definecolor{ruddy}{rgb}{1.0, 0.0, 0.16}
\definecolor{cadmiumred}{rgb}{0.89, 0.0, 0.13}
\definecolor{psychedelicpurple}{rgb}{0.87, 0.0, 1.0}
\pgfplotsset{every tick label/.append style={font=\tickSize}}
\def\tickSize{\scriptsize}
\def\labelSize{\scriptsize}
\def\PlotHeight{8.5em}
\def\rewPlotWidth{0.38\textwidth}
\def\violationsPlotWidth{0.38\textwidth}

\definecolor{aqua}{rgb}{0.0, 1.0, 1.0}
\definecolor{ultramarineblue}{rgb}{0.25, 0.4, 0.96}
\definecolor{royalblue}{rgb}{0.25, 0.41, 0.88}
\definecolor{aquamarine}{rgb}{0.5, 1.0, 0.83}
\definecolor{ballblue}{rgb}{0.13, 0.67, 0.8}
\colorlet{low_color}{aqua}
\colorlet{medium_color}{ballblue}
\colorlet{high_color}{blue}

%% file: 00_Abstract.tex
Safe exploration is a key challenge in Reinforcement Learning (RL) that aims to prevent agents from making decisions that could cause harm while exploring their environment.
 Shielding is one such technique that assumes domain knowledge in the form of a model of the environment to decide upon the safety of an action.
 Although shielding is a well-established approach, it has seen limited application in RL due to significant barriers.
 The main obstacle is the lack of an accessible end-to-end infrastructure that connects formal shield synthesis with standard RL frameworks.
 As a result, applying shielding 
 requires expertise in formal methods and substantial engineering effort to construct probabilistic models, which has kept it outside the typical RL workflow.
 We address this problem by extending our shield synthesis tool \tempest into a practical backend for safe RL.
 Our core contribution is \tempestpy, a Python library that integrates \tempest-based shield synthesis directly into the Gymnasium API, allowing shields to be synthesized and deployed within existing RL pipelines.
 This substantially lowers the barrier to entry for shielding and turns formal safe-exploration methods into a usable component for RL practitioners.
 In addition, we extend {\tempest}'s algorithmic support to compute sound shields for stochastic multiplayer games, thereby preserving formal safety guarantees.
 In this setting, specifying a model of the environment becomes the remaining requirement rather than the primary integration challenge.
 We demonstrate the resulting workflow end to end and evaluate shielded and unshielded RL across multiple environments.
 To facilitate the modeling step, we provide symbolic models for \minigrid and introduce \minigridsafe, a collection of playground environments designed to make shielding easily accessible and experimentally transparent.
 \minigridsafe extends MiniGrid with safety-oriented scenarios featuring probabilistic transitions and additional agents, enabling the study of challenging safety aspects in a simple and intuitive setting.

%% file: 01_Introduction.tex
Deep reinforcement learning (DRL) has achieved impressive results in complex decision-making tasks, ranging from games such as AlphaGo to applications in robotics and autonomous driving~\cite{mousavi2018deep}.
However, safety remains a central challenge for deploying DRL in real-world applications.
In DRL, an agent learns by exploring its environment through trial and error.
As a result, the agent may select actions that could potentially cause harm to itself or to its environment.
In recent years, \emph{shielding} has emerged as a widely studied theoretical approach to safe reinforcement learning~\cite{DBLP:conf/tacas/BloemKKW15}.
This technique introduces a logical component, called a shield, that monitors the environment and prevents the execution of actions that violate formally defined safety constraints.
This allows the agent to continue learning while maintaining safety throughout exploration.
Despite the theory of shielding and multiple extensions being well studied
, it has not seen a lot of applications in RL research yet.
To integrate a shield into an existing RL setting, several barriers must be overcome.
The first barrier is tool support for generating and integrating shields.
To the best of our knowledge, \tempest~\cite{DBLP:conf/atva/PrangerKPB21} is the only available shield-synthesis tool for probabilistic environments.
Built on top of \storm~\cite{hensel2021probabilistic}, a state-of-the-art probabilistic model checker implemented in C++, \tempest provides efficient algorithms for probabilistic analysis and shield synthesis.
However, most experimental work in RL is conducted in Python, whereas \tempest is primarily used through command-line interfaces.
As a result, users must manually integrate shields into their RL frameworks.
This lack of seamless integration has made \tempest impractical for most RL workflows in practice.
The second barrier concerns modeling.
Shields are rigorously computed from a \emph{formal specification} of safety-critical properties and a \emph{symbolic model} of the environment, typically in the form of a Markov decision process (MDP) or a stochastic multiplayer game (SMG).
Constructing such a model can require substantial effort.
In particular, it requires expertise in symbolic modeling languages, such as the PRISM language~\cite{KNP11}, which many RL practitioners do not possess.
Moreover, realistic environments often contain details that are irrelevant to the safety property under consideration.
As a result, it is often necessary to derive abstractions that preserve the relevant safety guarantees, which in turn requires considerable expertise in symbolic modeling.
To alleviate these barriers, this paper makes three contributions.
First, we present \tempestpy, a Python interface for \tempest that enables shields to be integrated into standard RL frameworks.
Second, we extend \minigrid, a popular library for RL environments, with features that are relevant for safe RL and provide a complete shielding toolchain for the resulting environments.
For this extension, called \minigridsafe, we provide automatic translations to symbolic \prism models from the environment definition, giving practitioners legible, concrete examples that make symbolic modeling more accessible.
Third, we extend \tempest with sound algorithms for shield computation in stochastic multiplayer games.
As the core of our contribution, we present \tempestpy, a Python library built on top of \tempest.
Using \tempestpy, shielding can be incorporated directly into the standard Gymnasium API, which is used by many popular RL frameworks.
This allows users to synthesize shields directly from Python and deploy them through standard environment wrappers.
Previously, users had to manually parse the lookup table generated by \tempest and integrate it into their RL framework.
\tempestpy eliminates this step by exposing shield synthesis and application through a Python interface.
To lower the modeling barrier, we provide a complete toolchain for \minigridsafe.
While \minigrid is well suited for experimentation, it does not natively support features that are central to safe RL, such as probabilistic dynamics or multiple interacting agents.
We therefore extend the library with stochastic dynamics and additional agents with probabilistic or adversarial behavior.
For \minigridsafe environments, symbolic MDP and SMG models are generated automatically in the \prism modeling language and can be used directly as input to \tempestpy.
The generated models are human-readable, making the relationship between the RL environment and the symbolic model explicit to the RL practitioner.
We demonstrate the complete workflow for shielded RL by creating models, synthesizing shields, and integrating them into the Gymnasium API through standard environment wrappers.
Lastly, we extend the algorithmic capabilities of \tempest with sound implementations for reachability in SMGs.
With sound solution methods available for the computation of shields for MDPs and SMGs, \tempestpy automatically defaults to these sound computations. Thus, the generated shields provide the formal guarantees required by the formal safety specifications.
We conclude by providing shielding demonstrations and presenting learning results for both shielded and unshielded RL in the provided environments.

%% file: 02_RelatedWork.tex
\ph{Related Work}
%%Thus, the policy might contain unsafe behavior without the shield.
%%\pet employs partial-exploration techniques to speed up the verification process.
%%Using heuristics, only parts of the state space have to be explored, while ensuring correctness.
%%\pet does not provide a way to export strategies.
Shielding has been extended along several dimensions.
\cite{DBLP:conf/aaai/Carr0JT23} extend shields to partially observable environments,
\cite{brorholt2023shielded} computed shields for hybrid systems, and
\cite{DBLP:conf/aaai/RodriguezAC0K25} extend shield synthesis to richer specification logics by supporting LTL modulo theories,
enabling shields for systems with infinite or continuous domains.
\cite{DBLP:conf/ifaamas/BrorholtL025} address scalability in multi-agent systems via compositional assume-guarantee reasoning.
\cite{DBLP:conf/nips/BanerjeeRBD24} replace fixed backup policies with a dynamic planner that jointly optimises safety and task progress and
\cite{DBLP:conf/rlc/CorsiARK0F24} combine neural network verification with shielding,
activating the shield only in regions of the state space identified as unsafe.
A broader survey of shielding approaches and their trade-offs is given in~\cite{DBLP:journals/cacm/KonighoferBJJP25}.
Tools that connect formal methods with RL, such as COOL-MC~\cite{gross2022cool} and MoGym~\cite{gros2022mogym}
focus on policy verification rather than runtime enforcement.
\cite{DBLP:conf/icaart/GrossS24} extend this line of work,
verifying stochastic RL policies
against PCTL specifications using \storm.
Despite the advances in the theory of shielding, accessible end-to-end infrastructure for deploying shields within standard RL workflows remains scarce, a gap that \tempestpy directly addresses.
The underlying engines for shield synthesis are probabilistic model checkers.
\storm~\cite{hensel2021probabilistic} and its Python bindings \stormpy serve as the backend for \tempestpy,
PRISM~\cite{KNP11} and PRISM-games~\cite{DBLP:conf/cav/KwiatkowskaN0S20} support strategy synthesis, 
but have no native support for exporting the non-deterministic strategies required for shielding.
PET~\cite{meggendorfer2024pet} further extends game-based analysis to stochastic games.

%% file: 03_Background.tex
A \emph{Markov decision process (MDP)} is a tuple $\mdp = \langle \states, s_0, \Act, \trans \rangle$ where $\states$ is a finite set of states, $s_0 \in \states$ is the initial state, $\Act$ is a finite set of actions, and $\trans : \states \times \Act \rightarrow \Dist(\states)$ is the probabilistic transition function.
For all $s \in \states$, the available actions are $\Act(s) = \{a \in \Act \mid \exists s' \in \states.\ \trans(s, a)(s') \neq 0\}$, and we assume $|\Act(s)| \geq 1$.
A \emph{reward function} $\rewFunction \colon \states \times \Act \rightarrow \mathbb{R}$ assigns a reward to each state-action pair.
A \emph{stochastic multiplayer game (SMG)} is a tuple $\smg = \langle \players, \states, s_0, \Act, \trans, (\states_i)_{i \in \players} \rangle$ where $\players$ is a finite set of players, $\states$ is a finite set of states, $s_0 \in \states$ is the initial state, $\Act$ is a finite set of actions, $\trans : \states \times \Act \rightarrow \Dist(\states)$ is the probabilistic transition function, and $(\states_i)_{i \in \players}$ is a partition of $\states$, i.e., $\states = \bigcup_{i \in \players} \states_i$ and $\states_i \cap \states_j = \emptyset$ for all $i \neq j$.
For all $s \in \states$, the available actions are $\Act(s) = \{a \in \Act \mid \exists s' \in \states.\ \trans(s,a)(s') \neq 0\}$, and we assume $|\Act(s)| \geq 1$.
If $s \in \states_i$, then player $i$ chooses the action to be taken in state $s$.
\ph{Probabilistic Model Checking}
We consider safety properties expressed in \emph{Probabilistic Computation Tree Logic (PCTL)}~\cite{BK08} for MDPs and \emph{Reactive Modules Probabilistic Alternating-time Temporal Logic (RPATL)}~\cite{DBLP:conf/tacas/ChenFKPS12} for SMGs, which extends PCTL by allowing properties to be quantified over coalitions of players.
In the following, we use $\varphi$ to denote a safety formula in either PCTL or RPATL, depending on the underlying model.
Probabilistic model checking~\cite{BK08} computes the probability of satisfying $\varphi$ over a finite or infinite horizon.
We denote by $\horiz \in \mathbb{N}$ the horizon encoded in $\varphi$; for the unbounded case, $\horiz = \infty$.
For a given MDP $\mdp$ and a safety property $\varphi$, model checking computes the following:
\begin{itemize}[leftmargin=1.2em]
    \item $\mathbb{P}_{\mdp^\pi, \varphi} \colon \mathcal{S} \times \mathbb{N} \rightarrow [0,1]$ is the probability of satisfying $\varphi$ in the Markov chain $\mdp^\pi$ induced by a policy $\pi$, starting from a state $s \in \mathcal{S}$ within $\horiz$ steps.
    \item $\mathbb{P}^{\mathsf{max}}_{\mdp, \varphi}(s,\horiz) = \max_{\pi}\, \mathbb{P}_{\mdp^\pi, \varphi}(s,\horiz)$ is the \emph{maximal} probability of satisfying $\varphi$ from state $s$ within $\horiz$ steps, maximised over all policies.
    \item $\mathbb{P}^{\mathsf{min}}_{\mdp, \varphi}(s,\horiz) = \min_{\pi}\, \mathbb{P}_{\mdp^\pi, \varphi}(s,\horiz)$ is the \emph{minimal} probability of satisfying $\varphi$ from state $s$ within $\horiz$ steps, minimised over all policies.
\end{itemize}
\noindent For an SMG $\smg$, $\players$ is partitioned into two coalitions: the \emph{maximising coalition} $\players_{\max} \subseteq \players$, whose joint policy $\pi_{\max}$ seeks to satisfy $\varphi$, and the \emph{minimising coalition} $\players_{\min} = \players \setminus \players_{\max}$, whose joint policy $\pi_{\min}$ acts adversarially.
A strategy profile $\boldsymbol{\pi} = (\pi_{\max}, \pi_{\min})$ induces a Markov chain $\smg^{\boldsymbol{\pi}}$, and model checking generalises to the following quantities:
\begin{itemize}[leftmargin=1.2em]
    \item $\mathbb{P}_{\smg^{\boldsymbol{\pi}}, \varphi} \colon \mathcal{S} \times \mathbb{N} \rightarrow [0,1]$ is the probability of satisfying $\varphi$ in the Markov chain $\smg^{\boldsymbol{\pi}}$ induced by a strategy profile $\boldsymbol{\pi} = (\pi_{\max}, \pi_{\min})$, starting from state $s \in \mathcal{S}$ within $\horiz$ steps.
\item $\mathbb{P}^{\mathsf{max}}_{\smg, \varphi}(s,\horiz) =
\max_{\pi_{\max}}\min_{\pi_{\min}}\,
\mathbb{P}_{\smg^{\boldsymbol{\pi}}, \varphi}(s,\horiz)$
is the \emph{maximal} probability of $\varphi$ being satisfied from state $s$ within $\horiz$ steps, maximised over $\pi_{\max}$ against a player $\pi_{\min}$ that minimises the probability of satisfaction.
\item $\mathbb{P}^{\mathsf{min}}_{\smg, \varphi}(s,\horiz) =
\min_{\pi_{\min}}\max_{\pi_{\max}}\,
\mathbb{P}_{\smg^{\boldsymbol{\pi}}, \varphi}(s,\horiz)$
is the \emph{minimal} probability of $\varphi$ being satisfied from state $s$ within $\horiz$ steps, minimised over $\pi_{\min}$ against a player $\pi_{\max}$ that maximises the probability of satisfaction.
\end{itemize}
\ph{Reinforcement Learning}
In reinforcement learning (RL)~\cite{sutton2018reinforcement}, an agent learns a task via interactions with an unknown environment modelled by an MDP $\mdp = \langle \states, s_0, \Act, \trans, \rewFunction \rangle$.
In each state $s \in \states$, the agent chooses an action $a \in \Act$, and the environment transitions to a successor state $s'$ with probability $\trans(s, a)(s')$.
An RL agent seeks to learn a \emph{policy} $\pi(a|s)$ that maximises the expected return, expressed as the discounted cumulative reward $\mathbb{E}\!\left[\sum_{t=0}^{\infty} \gamma^t \rewFunction(s_t, a_t)\right]$ with discount factor $\gamma \in [0,1)$.

%% file: 04a_Method_Synthesis.tex
In this section, we present \tempestpy, our Python library built on top of \tempest.
The shielding workflow is divided into an \emph{offline synthesis phase},
in which the shield is computed from a symbolic model and a safety property,
and an \emph{online application phase}, in which the shield is enforced at runtime during RL training or deployment.
\subsection{Shield Synthesis}\label{subsec:synthesis}
\begin{figure}[t]
    \centering
    \input{figures/offline_part.tex}
    \caption{The shield synthesis procedure. Dashed boxes indicate inputs.}
    \label{fig:offline_synthesis_part}
\end{figure}
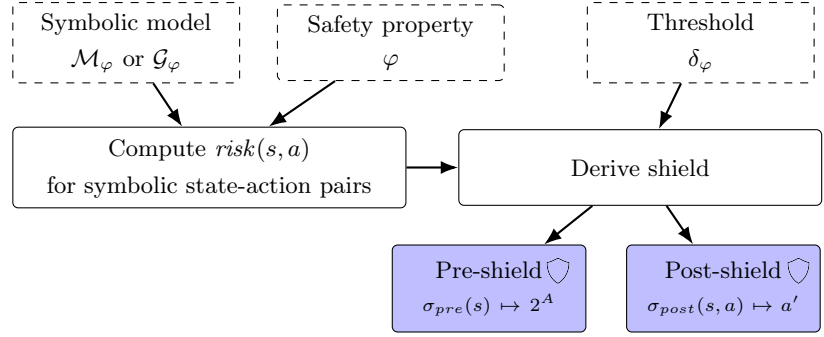

The essential property of a shield is its ability to provide provable safety guarantees during the learning and deployment of an RL agent.
To this end, shields are rigorously computed from a safety-critical abstraction of the RL environment and a formal safety property.
Figure~\ref{fig:offline_synthesis_part} illustrates the synthesis procedure.
\ph{Symbolic Models}
We assume that the RL problem at hand can be modelled as an MDP $\mdp$, a standard assumption in RL.
To synthesize a shield, one first derives a \emph{safety-critical abstraction} $\mdp_\varphi$ from the original MDP by retaining only those state features relevant to the specification $\varphi$.
By disregarding features irrelevant to safety, $\mdp_\varphi$ can be substantially smaller than $\mdp$, making synthesis tractable.
To obtain worst-case safety guarantees, probabilistic transitions can further be replaced by adversarial choices, yielding an SMG $\smg_\varphi$ as the underlying model for shield computation.
Both model types are supported by \tempestpy and must be expressed in the \prism modelling language.
Computing $\mdp_\varphi$ or $\smg_\varphi$ is problem-dependent and can require considerable effort; in Section~\ref{sec:minigridsafe}, we show how this step is automated for \minigridsafe environments.
\ph{Safety Properties and Shielding Objectives}
To synthesize a shield, the user must specify the safety property and the probability threshold the shield is required to enforce.
Given an abstract MDP $\mdp_\varphi$, the user provides a \emph{shielding objective} \mdpobjective, where $\varphi$ is a safety property in PCTL and $\thresh \in [0,1]$ is a probability threshold.
For an abstract SMG $\smg_\varphi$, the safety property $\varphi$ is expressed in RPATL.
This formula encodes the coalition $\coali \subseteq \players$ of players whose actions are guarded by the shield.
In states controlled by players in $\players \setminus \coali$, actions are chosen \emph{adversarially}, yielding worst-case safety guarantees that hold regardless of environment behaviour.
\ph{Shield Computation}
In the following, we assume that $\varphi$ specifies that no unsafe state is eventually reached.
Under this formulation, the \emph{risk} of a state-action pair $(s,a)$ is defined as the probability of eventually violating $\varphi$ when taking action $a$ in state $s$.
Using the safety-critical abstraction $\mdp_\varphi$ or $\smg_\varphi$ and the shielding objective \mdpobjective, \tempestpy computes the risk of taking each action $a$ in each state $s$.
The risk is obtained by calculating, for every state-action pair $(s,a)$, the probability of violating $\varphi$ when executing $a$ and subsequently following an optimal policy.
An action $a$ in state $s$ is classified as \emph{safe} if its risk does not exceed the threshold $\thresh$, and as \emph{unsafe} otherwise.
\begin{definition}[State-Action Risk]\label{def:sa_risk}
Given an abstract MDP $\mdp_\varphi$ or SMG $\smg_\varphi$ and a safety property $\varphi$, the risk function $\risk_{\phi} : \states \times \Act \rightarrow [0,1]$ is defined as:
$$\forall s \in \states,\, \forall a \in \Act:\quad \risk_{\phi}(s,a) = \sum_{s'\in \states} \trans(s,a)(s')\cdot \mathbb{P}^{\mathsf{min}}_{\mdp_\varphi, \varphi}(s', \horiz-1),$$
where $\horiz$ denotes the horizon bound encoded in $\varphi$.
For unbounded properties, $\horiz = \infty$.
\end{definition}
\noindent With the risk of each state-action pair $(s,a)$ computed, we now discuss how the different types of shields are synthesized.
A \emph{pre-shield} $\presh : \states \rightarrow 2^{\Act}$ acts \emph{before} the agent selects an action.
Given the current state $s$, it returns the set of admissible actions, i.e., those whose risk does not exceed $\thresh$:
$$\presh(s) = \{a \in \Act(s) \mid \risk_{\phi}(s,a) \leq \thresh\}.$$
The agent is then restricted to select only from $\presh(s)$, preventing unsafe actions from ever being executed.
If the agent is in a state where no action satisfies the risk threshold, i.e., $\presh(s) = \emptyset$, we call this state \emph{dangerous} and the shield falls back to the set of least unsafe actions:
$$\presh(s) = \{a \in \Act(s) \mid \risk_{\phi}(s,a) = \min_{a' \in \Act(s)}\, \risk_{\phi}(s,a')\}.$$
A \emph{post-shield} $\postsh : \states \times \Act \rightarrow \Act$ acts \emph{after} the agent has selected an action.
Given the current state $s$ and the agent's chosen action $a$, it returns a safe action to execute:
$$\postsh(s, a) = \begin{cases} a & \text{if } \risk_{\phi}(s,a) \leq \thresh, \\ f(s,a) & \text{otherwise,}\end{cases}$$
where $f : \states \times \Act \rightarrow \Act$ is a user-defined fallback function that selects a replacement action from the safe actions available in $s$.
The post-shield thus intervenes only when the agent's chosen action is unsafe, leaving the agent's behaviour unaffected otherwise.
The choice of $f$ is left to the user. 
A simple choice is a fixed, safe action $a_{\mathit{safe}} \in \Act$, such as a no-op or a braking action.
Alternatively, $f$ can select the safest available action $\arg\min_{a \in \presh(s)}\, \risk_{\phi}(s,a)$,
or $f$ can be extended to return the safe action with the highest Q-value, to minimise disruption to the agent's learned policy.

%% file: figures/offline_part.tex
\begin{tikzpicture}[
  font=\small,
  >=Latex,
  input/.style={
    draw,
    dashed,
    rounded corners=2pt,
    align=center,
    minimum width=3.0cm,
    minimum height=1.0cm,
    inner sep=4pt
  },
  block/.style={
    draw,
    rounded corners=2pt,
    align=center,
    minimum width=3.4cm,
    minimum height=1.0cm,
    inner sep=4pt
  },
  artifact/.style={
    draw,
    rounded corners=2pt,
    align=center,
    minimum width=3.2cm,
    minimum height=1.15cm,
    inner sep=4pt
  },
  arrow/.style={
    -Latex,
    thick
  },
  phase/.style={
    draw,
    rounded corners=4pt,
    thick,
    inner sep=8pt
  }
]
\node[input] (model) at (-2.3, 2.7)
{Symbolic model\\[1mm] $\mdp_\varphi$ or $\smg_\varphi$};
\node[input] (property) at (1.2, 2.7)
  {Safety property\\[0.5mm] $\varphi$};
\node[input] (threshold) at (5.3, 2.7)
  {Threshold\\[1mm] $\thresh$};
\node[block, minimum width=5.2cm] (risk) at (-1.2, 1.05)
  {Compute $\risk(s,a)$\\[1mm]
   for symbolic state-action pairs};
\node[block, minimum width=4.8cm] (derive) at (4.5, 1.05)
  {Derive shield};
\node[
  artifact,
  minimum width=2.55cm,
  minimum height=1.15cm,
  inner sep=3pt,
  fill=blue!25,
  text width=2.35cm
] (preshield) at (2.5, -0.55)
  {
    Pre-shield\\[0.5mm]
    {\scriptsize $\presh(s) \mapsto 2^A$}
  };
\node[inner sep=0pt] at ($(preshield.north east)+(-0.28,-0.26)$)
  {\includegraphics[bb=15 10 800 780,scale=0.018]{pictures/shield-icon-blank.png}};
\node[
  artifact,
  minimum width=2.55cm,
  minimum height=1.15cm,
  inner sep=3pt,
  fill=blue!25,
  text width=2.35cm
] (postshield) at (5.6, -0.55)
  {
    Post-shield\\[0.5mm]
    {\scriptsize $\postsh(s,a) \mapsto a'$}
  };
\node[inner sep=0pt] at ($(postshield.north east)+(-0.20,-0.26)$)
  {\includegraphics[bb=15 10 800 780,scale=0.018]{pictures/shield-icon-blank.png}};
\draw[arrow] (model) -- (risk);
\draw[arrow] (property) -- (risk);
\draw[arrow] (risk) -- (derive);
\draw[arrow] (threshold) -- (derive);
\draw[arrow] (derive) -- (preshield);
\draw[arrow] (derive) -- (postshield);
\end{tikzpicture}

%% file: 04b_Method_Application_RL.tex
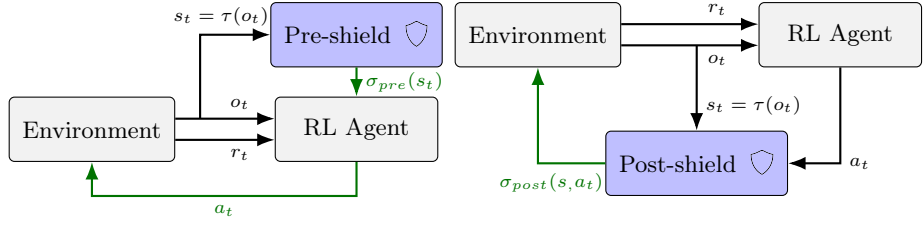
\begin{figure}[t]
    \centering
    \input{figures/online_shielding_part.tex}
    \caption{The two types of shield application. \textit{Left:} Pre-shielding restricts the choices of the agent. \textit{Right:} Post-shielding prevents unsafe actions from being executed.}
    \label{fig:online_shielding_part}
\end{figure}
\subsection{Shield Application}
Integrating a synthesized shield into an RL training loop requires two ingredients:
a \emph{translation function} $\rlTrans$ that maps RL environment observations to states of the abstract model,
and a \emph{shield wrapper} that enforces the shield at runtime.
\tempestpy provides two wrappers built on top of the Gymnasium interface~\cite{kwiatkowski2024gymnasium}, \texttt{PreShieldWrapper} and \texttt{PostShieldWrapper}.
In the following, we discuss how \tempestpy synthesizes the shield and how the two types of shields are applied to the RL algorithm.
Figure~\ref{fig:online_shielding_part} provides illustrations of pre- and post-shielding.
\ph{Shield Factory}
In the offline synthesis phase, \tempestpy computes the risk values $\risk_\phi(s,a)$ for all state-action pairs of $\mdp_\varphi$ or $\smg_\varphi$ under the shielding objective $\mdpobjective$, as described in Section~\ref{subsec:synthesis}.
Since this computation can be intensive, it is performed once prior to training and the resulting risk values are stored.
In \tempestpy, this process is encapsulated in a \emph{shield factory}, which accepts the \prism model and the safety property $\varphi$ as inputs and internally runs the associated model checking query.
Safety properties are expressed as \prism query strings, following the property specification languages of \prism~\cite{KNP11} and \prismgames~\cite{DBLP:conf/cav/KwiatkowskaN0S20}.
To derive a shield from the computed risk values, the user provides the threshold \thresh, classifying each state-action pair as either safe or unsafe.
Listing~\ref{lst:shield-build} shows how a shield is constructed.
\begin{lstlisting}[
  float=ht,
  language=python, mathescape,
  basicstyle=\fontsize{8.5}{6}\selectfont\ttfamily,
  numbers=none,
  frame=single, breaklines=true,
  captionpos=b,
  caption={Constructing a shield using \tempestpy.},
  label={lst:shield-build}]
factory = ShieldFactory(
    model="symbolic_mdp.prism",
    property="Pmin=? [ F<=15 \"unsafe\" ]",
)
shield = factory.build(ShieldConfig(threshold=0.05))
\end{lstlisting}
\vspace{-2em}
\ph{Observation Translation}
The translation function $\rlTrans$ bridges the RL environment and the abstract model $\mdp_\varphi$ by mapping each environment observation to a full assignment of \prism variables, identifying the corresponding abstract state $s \in \states$.
The user provides $\rlTrans$ as a Python function whose return value is a dictionary mapping \prism variable names to their current values.
The concrete form of $\rlTrans$ is determined by the choice of abstraction: since $\mdp_\varphi$, or $\smg_\varphi$, retains only the state features relevant to $\varphi$, the translation projects the observation onto this subset of features, discarding those irrelevant to safety.
In the simplest case, the retained features are read directly from the observation, as when the agent's coordinates coincide with the \prism state variables; when the abstraction is coarser, $\rlTrans$ must discretize continuous values according to the state space of $\mdp_\varphi$.
\ph{Pre-Shield Integration}
The pre-shield wrapper enforces $\presh$ by restricting the agent's actions before each decision.
At every step, the wrapper translates the current observation to the abstract state ${s}_t = \rlTrans(o_t)$ and computes the admissible action set $\presh({s}_t)$, which is exposed to the training algorithm as a Boolean action mask.
The agent therefore only ever selects actions from the safe set, and unsafe actions are never executed.
Pre-shielding requires a maskable RL algorithm that conditions its policy on the provided mask. 
In dangerous states, the wrapper automatically falls back to the set of least-unsafe actions as defined in Section~\ref{subsec:synthesis}.
\ph{Post-Shield Integration}
The post-shield wrapper intercepts the agent's chosen action after selection and replaces it with a safe alternative when the proposed action violates the shield constraint.
The agent observes the full, unmasked action space and selects freely; the shield acts as a transparent safety filter between the agent and the environment.
Post-shielding is therefore compatible with any RL algorithm, without requiring maskable implementations, at the cost of potentially introducing a bias in the learning signal when actions are corrected.

%% file: figures/online_shielding_part.tex
\begin{tikzpicture}[
  font=\small,
  >=Latex,
  box/.style={
    draw,
    rounded corners=2pt,
    align=center,
    minimum height=0.85cm,
    inner sep=5pt
  },
  envagent/.style={
    box,
    fill=gray!10,
    minimum width=2.15cm
  },
  shieldbox/.style={
    box,
    fill=blue!25,
    minimum width=2.0cm
  },
  arrow/.style={
    -Latex,
    thick
  },
  safearrow/.style={
    -Latex,
    thick,
    green!45!black
  },
  labeltext/.style={
    font=\scriptsize
  }
]
\node[envagent] (preenv) at (-5.6,0.10)
  {Environment};
\node[envagent] (preagent) at (-2.1,0.10)
  {RL Agent};
\node[shieldbox] (preshield) at (-2.1,1.35)
  {Pre-shield\hspace{5mm}};
\node[inner sep=0pt] at ($(preshield.east)+(-0.25,0.10)$)
  {\includegraphics[bb=15 10 800 780,scale=0.018]{pictures/shield-icon-blank.png}};
\draw[arrow]
  ($(preenv.east)+(0,0.14)$)
  -- node[above, pos=0.65, labeltext] {$o_t$}
     coordinate[pos=0.25] (preobsbranch)
  ($(preagent.west)+(0,0.14)$);
\draw[arrow]
  ($(preenv.east)+(0,-0.14)$)
  -- node[below, pos=0.65, labeltext] {$r_t$}
  ($(preagent.west)+(0,-0.14)$);
\draw[arrow]
  (preobsbranch)
  |- node[pos=0.65, above, labeltext] {${s}_t=\rlTrans(o_t)$}
  (preshield.west);
\draw[safearrow]
  (preshield.south)
  -- node[right, labeltext] {$\presh({s}_t)$}
  (preagent.north);
\draw[safearrow]
  (preagent.south)
  -- ++(0,-0.45)
  -| node[pos=0.25, below, labeltext] {$a_t$}
  (preenv.south);
\node[envagent] (postenv) at (0.3,1.35)
  {Environment};
\node[envagent] (postagent) at (4.3,1.35)
  {RL Agent};
\node[shieldbox] (postshield) at (2.4,-0.35)
  {Post-shield\hspace{5mm}};
\node[inner sep=0pt] at ($(postshield.east)+(-0.25,0.10)$)
  {\includegraphics[bb=15 10 800 780,scale=0.018]{pictures/shield-icon-blank.png}};
\draw[arrow]
  ($(postenv.east)+(0,0.14)$)
  -- node[above, pos=0.70, labeltext] {$r_t$}
  ($(postagent.west)+(0,0.14)$);
\draw[arrow]
  ($(postenv.east)+(0,-0.14)$)
  -- node[below, pos=0.70, labeltext] {$o_t$}
     coordinate[pos=0.55] (postobsbranch)
  ($(postagent.west)+(0,-0.14)$);
\draw[arrow]
  (postobsbranch)
  -- node[pos=0.70, right, labeltext] {${s}_t=\rlTrans(o_t)$}
  (postshield.north);
\draw[arrow]
  (postagent.south)
  |- node[right, labeltext] {$a_t$}
  (postshield.east);
\draw[safearrow]
  (postshield.west)
  -| node[pos=0.40, below, labeltext] {$\postsh(s_,a_t)$}
  (postenv.south);
\end{tikzpicture}

%% file: 06_MinigridSafe.tex
By using \tempestpy, shields can be easily integrated into RL as described in the previous section.
The second barrier, the construction of faithful symbolic models of RL environments, remains a formidable challenge.
In order to address this barrier, we introduce \minigridsafe as an extension of the popular \minigrid RL library.
In a \minigridsafe environment agents traverse a 2D grid by moving in the cardinal directions.
\minigridsafe builds on this widely used class of RL environments by introducing several safety-critical features and providing an automated translation to PRISM.
This yields two concrete advantages for practitioners: (1) symbolic models for shield synthesis are generated automatically in a human-readable form, and (2) the environment and its symbolic model are aligned by construction, eliminating the need for a translation function \rlTrans.
In the following, we discuss the extensions introduced in \minigridsafe.
\input{pictures/minigridsafe_envs.tex}
\ph{Probabilistic Behaviour}
As a first extension, \minigridsafe adds two distinct types of probabilistic behaviour for the agent.
Slippery tiles may cause the agent to be displaced to adjacent tiles, where the displacement distribution can be specified by the user.
The example shown in Figure~\ref{subfig:cliff} uses tiles with direction-dependent transition probabilities:
on the blue tiles, the agent may slip to the left or right of its intended cardinal movement, with an additional fixed probability of slipping in the tile’s tilt direction, indicated by the arrow, independently of the chosen action.
A second form of probabilistic behaviour is introduced by \emph{sticky actions}: rather than executing a single action, the agent may repeat the same action multiple times with some probability, modelling control noise.
\ph{Adversarial Actors}
\minigridsafe supports the inclusion of additional actors in the environment, enabling users to model safety-critical scenarios involving interactions with independently acting entities, such as adversaries or other policy-driven agents.
Their behavior is specified by a sequence of tasks, such as \texttt{GoTo(x,y)}, \texttt{MoveRandom(n)}, \texttt{DoNothing(n)}, or \texttt{FollowAgent("red")}.
Each actor executes its current task until completion before proceeding to the next, allowing users to specify structured, stochastic, or adversarial behavior compactly.
The adversary shown in Figure~\ref{subfig:adversaries} follows a fixed patrol route encoded as a task sequence.
\input{listings/prism_example_listing.tex}
\ph{Automatic Translation to \prism}
For a given \minigridsafe environment, the library automatically generates the corresponding model in the \prism language.
As illustrated in Listing~\ref{lst:translated_prism}, the generated models remain human-readable: grid regions, movement constraints, and terminal conditions are expressed as named \prism formulae, and probabilistic transitions appear as explicit guarded commands, as seen in the slippery tile behaviour of \texttt{Agent\_Move\_North}, making the model straightforward to inspect and relate back to the original environment.
If adversarial actors are present within the environment, they are assumed to behave adversarially, resulting in the environment being encoded as an SMG $\smg_\varphi$, otherwise the environment is encoded as an MDP $\mdp_\varphi$.

%% file: pictures/minigridsafe_envs.tex
\begin{figure}[t!]
    \centering
    \begin{subfigure}[b]{0.35\textwidth}
        \centering
        \captionsetup{width=1.3\linewidth}
        \includegraphics[width=\textwidth]{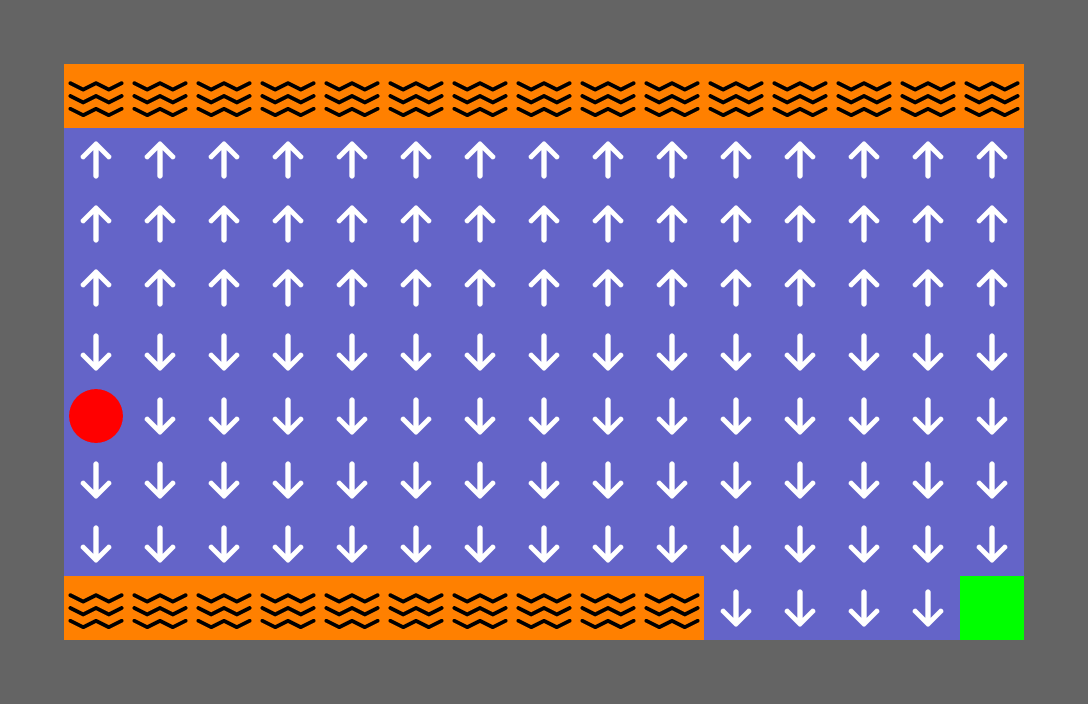}
        \vspace{0.2em}
        \caption{Environment with slippery tiles.}
        \label{subfig:cliff}
    \end{subfigure}
    \quad\quad\quad\quad\quad\quad
    \begin{subfigure}[b]{0.35\textwidth}
        \centering
        \captionsetup{width=1.5\linewidth}
        \input{pictures/adversary_annotated.tex}
        \vspace{0.3em}
        \caption{Environment with a patrolling adversary.}
        \label{subfig:adversaries}
    \end{subfigure}
    \caption{Two environments illustrating the safety-critical extensions of \minigridsafe.}
    \label{fig:minigridsafe_envs}
\end{figure}

%% file: pictures/adversary_annotated.tex
{\scalefont{1.0}
\begin{tikzpicture}
    \node[anchor=south west,inner sep=0] (image) at (0,0)
        {\includegraphics[width=0.9\textwidth]{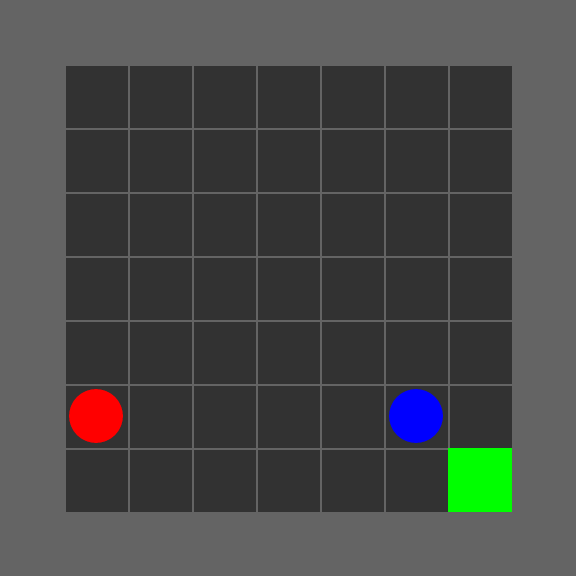}};
    \begin{scope}[x={(image.south east)},y={(image.north west)}]
        \draw [-stealth, line width=1.2pt, blue]
            (0.72, 0.34)   
            -- (0.72, 0.72) 
            -- (0.28, 0.72) 
            -- (0.28, 0.28) 
            -- (0.66, 0.28);
    \end{scope}
\end{tikzpicture}
}

%% file: listings/prism_example_listing.tex
\begin{lstlisting}[
  float=t,
  language=Matlab, mathescape,
  basicstyle=\fontsize{8.5}{6}\selectfont\ttfamily,
  numbers=none,
  frame=single, breaklines=true,
  caption={Snippets from the automatically translated \prism code.},
  captionpos=b, label={lst:translated_prism}
]
mdp

formula AgentCannotMoveNorth = (yAgent=1&xAgent>=1&xAgent<=14);
formula AgentCannotMoveEast = (xAgent=14&yAgent>=1&yAgent<=9);
[...]
formula AgentIsOnSlippery = (xAgent>=1&xAgent<=2&yAgent>=1&yAgent<=2) | (xAgent=13&yAgent=1) | (yAgent=2&xAgent>=13&xAgent<=14) | (xAgent>=1&xAgent<=14&yAgent>=3&yAgent<=9);
[...]
formula yAgentN = (AgentCannotMoveNorth ? yAgent : yAgent-1);
[...]
formula AgentIsInGoal = (xAgent=14&yAgent=1);
formula AgentIsInLava = (xAgent>=3&xAgent<=12&yAgent>=1&yAgent<=2);
formula AgentIsTerminal = AgentIsInGoal | AgentIsInLava | AgentDone;
[...]
module Agent
    xAgent : [1..14] init 1;
    yAgent : [1..9] init 1;

    [Agent_Move_North] !AgentIsTerminal & !AgentIsOnSlippery & !AgentCannotMoveNorth -> (yAgent'=yAgentN);
    [Agent_Move_East]  !AgentIsTerminal & !AgentIsOnSlippery & !AgentCannotMoveEast  -> (xAgent'=xAgentE);
    [Agent_Move_South] !AgentIsTerminal & !AgentIsOnSlippery & !AgentCannotMoveSouth -> (yAgent'=yAgentS);
    [Agent_Move_West]  !AgentIsTerminal & !AgentIsOnSlippery & !AgentCannotMoveWest  -> (xAgent'=xAgentW);
[...]
    [Agent_Move_North] !AgentIsTerminal & AgentIsOnSlippery -> 0.7 : (yAgent'=yAgentN) + 0.15 : (xAgent'=xAgentW) + 0.15 : (xAgent'=xAgentE);
[...]
endmodule

\end{lstlisting}

%% file: 07_Shielding_With_TempestPy.tex
\input{listings/preshielding_tempestpy_example.tex}
We now demonstrate the complete shielding workflow end to end using the two \minigridsafe environments introduced in Figure~\ref{fig:minigridsafe_envs}.
In the slippery cliff environment, Figure~\ref{subfig:cliff}, the agent must reach the goal while avoiding lava, formalized as the unbounded reachability property \texttt{Pmin=?\ [F\ "AgentLava"]}.
In the adversarial environment, Figure~\ref{subfig:adversaries}, the agent must avoid being caught by the adversary, formalized as \texttt{<<agent>>\ Pmin=?\ [F\ "AgentCaught"]}.
In the following, we compare training under the application of shields with varying thresholds and unshielded training.
We have used the implementations of MaskablePPO and PPO of Stable-Baselines3~\cite{SB3} for RL training.
For each threshold and unshielded training, we report the average over 5 runs.
\ph{Shielding Workflow}
The complete workflow requires three steps, all expressible within a standard Python training script.
First, the \prism model is generated automatically from the \minigridsafe environment via a single call to \texttt{write\_prism()}, producing a human-readable MDP or SMG encoding the environment.
The user is not required to write or inspect this model.
Second, a shield is synthesized from the generated model and the safety property using a \texttt{ShieldFactory}, which internally computes the risk of every state-action pair.
This computation is performed once, offline, prior to training.
Third, the environment is wrapped with a \texttt{PreShieldWrapper}, which exposes the shield as a Boolean action mask at every step of training, ensuring that unsafe actions are never executed.
Listing~\ref{lst:preshielding_workflow} shows the code that implements the three steps.
\ph{Results: Slippery Cliff}
In this environment, the agent has a fixed probability of $0.1$ of slipping north or south, depending on the tilt of the slippery tile.
The agent receives a reward of $1$ upon reaching the goal and a penalty of $-1$ when slipping into lava.
We train policies with two safety thresholds, $\thresh \in \{0.01, 0.001\}$, and without a shield.
Depending on the chosen threshold, the shield enforces a minimum distance to the lava tiles.
Results are shown in Figures~\ref{subfig:slippery_rewards} and~\ref{subfig:slippery_indicators}.
Both shields effectively prevent the agent from slipping into the lava, leading to high reward achieved early during training, while incuring only a small number of safety violations
Notably, the unshielded agent converges to a reasonable reward but continues to incur violations throughout training, demonstrating that task performance alone does not imply safety.
\ph{Results: Adversarial Environment}
In this environment, the agent has to navigate to the goal while avoiding being caught by a blue adversary.
The agent receives a reward of $1$ upon reaching the goal, $-1$ when caught by the adversary, and a fixed per-step penalty of $-0.01$.
We train policies with a shield that enforces safety with $\thresh = 0.0$ and one without.
Results are shown in Figures~\ref{subfig:adv_rewards} and~\ref{subfig:adv_indicators}.
The shielded agent quickly learns to reach the goal and is able to avoid the adversary persistently.
In contrast, the unshielded agent fails to learn an effective avoidance strategy, resulting in persistently low reward and a steadily increasing number of adversary catches throughout training.
This demonstrates the benefit of worst-case shield computation: by assuming adversary behavior, the agent is freed to focus on learning the task rather than discovering avoidance strategies through trial and error.
\input{plots/matrix.tex}

%% file: listings/preshielding_tempestpy_example.tex
\begin{lstlisting}[
  float=t,
  language=python, mathescape,
  basicstyle=\fontsize{8.5}{6}\selectfont\ttfamily,
  numbers=none,
  frame=single, breaklines=true,
  caption={Training a shielded Agent with a \texttt{PreShieldWrapper}.},
  captionpos=b, label={lst:preshielding_workflow}
]
# Build the shield from PRISM model and objective
env.unwrapped.write_prism(prism_path="abstract_model.prism")
factory = ShieldFactory(prism_path, property='Pmin=? [ F "AgentLava" ]')
config  = ShieldConfig(threshold=0.0)

# Wrap environment with PreShieldWrapper
env = PreShieldWrapper(env, factory, config,
    tau=lambda obs, info: {
        "xAgent": env.unwrapped.agent_pos[0],
        "yAgent": env.unwrapped.agent_pos[1]})

# Train
model = MaskablePPO("MlpPolicy", env)
model.learn(total_timesteps=500_000)
model.save("shielded_policy")
\end{lstlisting}

%% file: plots/matrix.tex
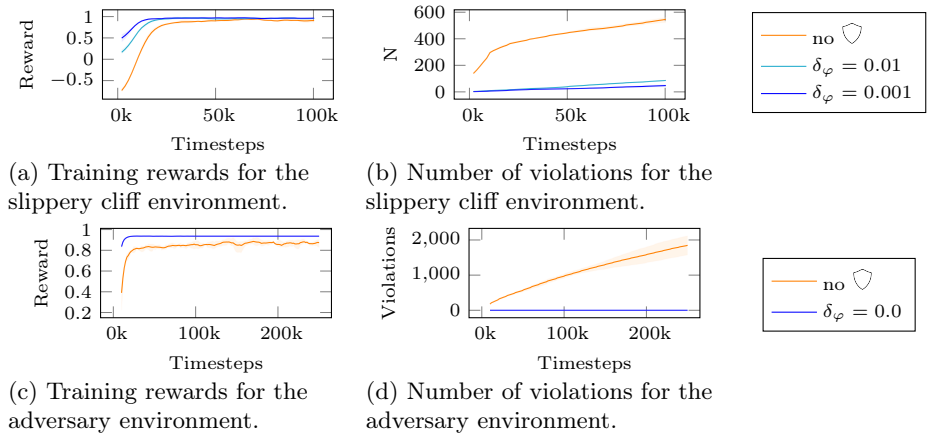
\begin{figure}[t]
  \begin{tabular}[c]{cc>{\centering\arraybackslash}p{0.25\textwidth}}
    \begin{subfigure}[t]{\rewPlotWidth}
      \centering
      \input{plots/lava_slippery_rewards.tex}
      \caption{Training rewards for the\\ slippery cliff environment.}
      \label{subfig:slippery_rewards}
    \end{subfigure}&
    \begin{subfigure}[t]{\violationsPlotWidth}
      \centering
      \input{plots/lava_slippery_violations.tex}
      \caption{Number of violations for the\\ slippery cliff environment.}
      \label{subfig:slippery_indicators}
    \end{subfigure}&
    \raisebox{2em}{
      \begin{tikzpicture}
        \begin{axis}[
          hide axis,
          xmin=0, xmax=1, ymin=0, ymax=1,
          legend style={draw=black, fill=none, font=\scriptsize},
          legend cell align={left},
        ]
          \addlegendimage{color=orange, solid}
          \addlegendentry{no \shieldIcon}
          \addlegendimage{color=medium_color, solid}
          \addlegendentry{$\delta_\varphi=0.01$}
          \addlegendimage{color=high_color, solid}
          \addlegendentry{$\delta_\varphi=0.001$}
        \end{axis}
      \end{tikzpicture}
    }\\
    \begin{subfigure}[t]{\rewPlotWidth}
      \centering
      \input{plots/adv_complex_rewards.tex}
      \caption{Training rewards for the\\ adversary environment.}
      \label{subfig:adv_rewards}
    \end{subfigure}&
    \begin{subfigure}[t]{\violationsPlotWidth}
      \centering
      \input{plots/adv_complex_violations.tex}
      \caption{Number of violations for the\\ adversary environment.}
      \label{subfig:adv_indicators}
    \end{subfigure}&
    \raisebox{2em}{
      \begin{tikzpicture}
        \begin{axis}[
          hide axis,
          xmin=0, xmax=1, ymin=0, ymax=1,
          legend style={draw=black, fill=none, font=\scriptsize},
          legend cell align={left},
        ]
          \addlegendimage{color=orange, solid}
          \addlegendentry{no \shieldIcon}
          \addlegendimage{color=high_color, solid}
          \addlegendentry{$\delta_\varphi=0.0$}
        \end{axis}
      \end{tikzpicture}
    }\\
  \end{tabular}
\caption{Training results for the environments in Figure~\ref{fig:minigridsafe_envs}. Thick lines show the mean over all runs and the shaded regions indicate the standard deviation.}
  \label{fig:matrix}
\end{figure}

%% file: plots/lava_slippery_rewards.tex
\begin{tikzpicture}
    \begin{axis}[
      label style={font=\labelSize},
      height=\PlotHeight,
      ylabel={Reward},
      xlabel={Timesteps},
      width=\textwidth,
      xticklabel style={
        /pgf/number format/fixed,
        /pgf/number format/precision=1
      },
      scaled ticks=false,
      xticklabel = {
        \pgfkeys{/pgf/fpu}
        \pgfmathparse{\tick/1000}
        \pgfmathprintnumber{\pgfmathresult}k
      },
    ]
    \rewardPlot{plots/data/lava_slippery_10_0.0.csv}{orange}
    \rewardPlot{plots/data/lava_slippery_10_0.99.csv}{medium_color}
    \rewardPlot{plots/data/lava_slippery_10_0.999.csv}{high_color}
    \end{axis}
  \end{tikzpicture}

%% file: plots/lava_slippery_violations.tex
\begin{tikzpicture}
    \begin{axis}[
      label style={font=\labelSize},
      ylabel={N},
      xlabel={Timesteps},
      width=\textwidth,
      height=\PlotHeight,
      xticklabel style={
        /pgf/number format/fixed,
        /pgf/number format/precision=1
      },
      scaled ticks=false,
      xticklabel = {
        \pgfkeys{/pgf/fpu}
        \pgfmathparse{\tick/1000}
        \pgfmathprintnumber{\pgfmathresult}k
      },
    ]
    \lavaPlot{plots/data/lava_slippery_10_0.0.csv}{orange}
    \lavaPlot{plots/data/lava_slippery_10_0.99.csv}{medium_color}
    \lavaPlot{plots/data/lava_slippery_10_0.999.csv}{high_color}
  \end{axis}
\end{tikzpicture}

%% file: plots/adv_complex_rewards.tex
\begin{tikzpicture}
    \begin{axis}[
      label style={font=\labelSize},
      height=\PlotHeight,
      ylabel={Reward},
      xlabel={Timesteps},
      xmax=265000,
      width=\textwidth,
      xticklabel style={
        /pgf/number format/fixed,
        /pgf/number format/precision=1
      },
      scaled ticks=false,
      xticklabel = {
        \pgfkeys{/pgf/fpu}
        \pgfmathparse{\tick/1000}
        \pgfmathprintnumber{\pgfmathresult}k
      },
    ]
    \rewardPlot{plots/data/adversary_unshielded.csv}{orange}
    \rewardPlot{plots/data/adversary_shielded.csv}{high_color}
    \end{axis}
  \end{tikzpicture}

%% file: plots/adv_complex_violations.tex
\begin{tikzpicture}
    \begin{axis}[
      label style={font=\labelSize},
      ylabel={Violations},
      xlabel={Timesteps},
      xmax=265000,
      width=\textwidth,
      height=\PlotHeight,
      yticklabel style={
        /pgf/number format/fixed,
        /pgf/number format/precision=1
      },
      scaled ticks=false,
      xticklabel = {
        \pgfkeys{/pgf/fpu}
        \pgfmathparse{\tick/1000}
        \pgfmathprintnumber{\pgfmathresult}k
      },
    ]
    \collisionPlot{plots/data/adversary_unshielded.csv}{orange}
    \collisionPlot{plots/data/adversary_shielded.csv}{high_color}
  \end{axis}
\end{tikzpicture}

%% file: 08_Tool_Architecture.tex
This work directly addresses the long-standing difficulty of integrating \tempest into Python-based RL workflows: previously, users had to manually parse \tempest's output, manage C++ dependencies, and write custom integration code for each RL framework.
\tempestpy solves this by letting users integrate shield synthesis fully into Python.
\tempestpy is implemented as an extension of \stormpy.
While \stormpy provides general-purpose Python bindings for \storm, \tempestpy is a shielding-specific extension that exposes risk computation, shield synthesis, and Gymnasium integration directly through a Python API.
Both \tempest and \tempestpy are maintained as forks of their respective upstreams and kept in sync through regular merges.
The tools are publicly available alongside documentation and examples covering the full shielding workflow, including shield synthesis and Gymnasium integration.\footnote{\url{https://tempest-synthesis.org/}}
\ph{Sound Shield Computation for SMGs}
A key algorithmic contribution of this work is the implementation of sound solution methods for stochastic multiplayer games within \tempest.
Prior to this work, Tempest supported shield synthesis for MDPs but lacked sound algorithms for SMGs.
We implement sound procedures for safety-properties in SMGs based on~\cite{eisentraut2022value}. 
The resulting shields carry the same formal safety guarantees as their MDP counterparts: the computed risk values are correct with respect to the worst-case adversarial coalition, and the synthesized shield provably enforces the specified safety threshold regardless of environment behaviour.
With the addition of sound algorithms for SMGs, \tempestpy automatically selects sound computation for both model types.

%% file: 09_Conclusion.tex
In this work, we presented \tempestpy, a Python library that integrates shield synthesis directly into the Gymnasium API, making shielded RL accessible within standard RL frameworks.
We further extended \tempest with sound algorithms for shield computation in stochastic multiplayer games.
Both tools are maintained as forks of \storm and \stormpy, which are kept up to date through regular synchronisation with their upstreams.
To lower the barrier of entry for shielding, we presented \minigridsafe: a library of safety-critical RL environments featuring probabilistic dynamics and adversarial agents, with automated translation to PRISM models, providing a transparent starting point for practitioners exploring shielded RL.
For future work, we are going to address a limitation of the current shielding approaches implemented in \tempest.
Currently, all model checking queries for shield synthesis operate on explicit-state models, restricting the applicability of shielded RL to small state spaces.
We aim to address this by extending \tempestpy with shield synthesis based on symbolic model checking.
Furthermore, we want to provide more automated model translations for RL libraries such as Isaac Lab for robotics and PettingZoo's multi-agent environments. 